\documentclass[twocolumn,letterpaper]{IEEEAerospaceCLS}  

\usepackage[]{graphicx}    
\newcommand{\ignore}[1]{}  

\usepackage[utf8]{inputenc} 
\usepackage[T1]{fontenc}    
\usepackage{latexsym, amssymb, amsfonts, amsmath, amstext}
\usepackage{url}            
\usepackage{amsfonts}       
\usepackage{nicefrac}       
\usepackage{microtype}
\usepackage{tabularx, booktabs}
\usepackage{makecell}
\usepackage{multirow}
\usepackage{cite}
\usepackage{bm, soul}

\begin{document}
\title{SPEED+: Next-Generation Dataset for Spacecraft Pose Estimation across Domain Gap}

\author{%
Tae Ha Park\\ 
Department of Aeronautics \& Astronautics\\
Stanford University\\
496 Lomita Mall, Stanford, CA 94305\\
tpark94@stanford.edu
\and 
Marcus M{\"a}rtens\\
Advanced Concepts Team\\
European Space Agency\\
2201 AZ Noordwijk, The Netherlands\\
marcus.maertens@esa.int
\and
Gurvan Lecuyer\\
Advanced Concepts Team\\
European Space Agency\\
2201 AZ Noordwijk, The Netherlands\\
gurvan.lecuyer@esa.int
\and
Dario Izzo\\
Advanced Concepts Team\\
European Space Agency\\
2201 AZ Noordwijk, The Netherlands\\
dario.izzo@esa.int
\and
Simone D'Amico\\ 
Department of Aeronautics \& Astronautics\\
Stanford University\\
496 Lomita Mall, Stanford, CA 94305\\
damicos@stanford.edu
\thanks{\footnotesize 978-1-6654-3760-8/22/$\$31.00$ \copyright2022 IEEE}              
}

\maketitle

\thispagestyle{plain}
\pagestyle{plain}

\begin{abstract}
Autonomous vision-based spaceborne navigation is an enabling technology for future on-orbit servicing and space logistics missions. While computer vision in general has benefited from Machine Learning (ML), training and validating spaceborne ML models are extremely challenging due to the impracticality of acquiring a large-scale labeled dataset of images of the intended target in the space environment. Existing datasets, such as Spacecraft PosE Estimation Dataset (SPEED), have so far mostly relied on synthetic images for both training and validation, which are easy to mass-produce but fail to resemble the visual features and illumination variability inherent to the target spaceborne images. In order to bridge the gap between the current practices and the intended applications in future space missions, this paper introduces SPEED+: the next generation spacecraft pose estimation dataset with specific emphasis on domain gap. In addition to 60,000 synthetic images for training, SPEED+ includes 9,531 hardware-in-the-loop images of a spacecraft mockup model captured from the Testbed for Rendezvous and Optical Navigation (TRON) facility. TRON is a first-of-a-kind robotic testbed capable of capturing an arbitrary number of target images with accurate and maximally diverse pose labels and high-fidelity spaceborne illumination conditions. SPEED+ is used in the second international Satellite Pose Estimation Challenge co-hosted by SLAB and the Advanced Concepts Team of the European Space Agency to evaluate and compare the robustness of spaceborne ML models trained on synthetic images.
\end{abstract} 

\tableofcontents

\section{Introduction}

Autonomous navigation about a noncooperative Resident Space Object (RSO) is an enabling technology for future in-orbit servicing and space logistics missions, such as refueling space assets \cite{restore_L} and active debris removal \cite{removedebris}. As these missions involve interaction with a noncooperative target at close proximity, accurate determination and tracking of the target's pose (i.e., position and orientation) with respect to the servicer spacecraft is vital for safe docking and capture. Moreover, driven by the limited on-board power and computational capabilities, a monocular camera is a favored choice of sensor as opposed to more complex systems such as stereovision, Range Detection and Ranging (RADAR) or Light Detection and Ranging (LIDAR). Recently, Machine Learning (ML) techniques based on Convolutional Neural Networks (CNN) have been explored in applications of vision-based spacecraft pose estimation and navigation \cite{Sharma2019AAS, Park2019AAS, Kisantal2020SPEC, Chen2019SatellitePE, PasqualettoCassinis2020CNNEKF, Proenca2019Photorealistic, Cassinis2021ORGL, Black2021PoseEstimationCygnus}; however, it is impractical to acquire large-scale imagery of the interested target spacecraft with accurate pose labels in space to train CNN models. This is the main reason that CNN models have not been deployed in space so far. In fact, spaceborne ML models up to date have been trained and, most importantly, validated virtually exclusively on synthetic imagery. While synthetic imagery is easy to mass-produce and annotate for training, it is prone to performance degradation when tested on the target spaeborne imagery as CNNs overfit to the features specific to the synthetic imagery. The phenomenon is known as \emph{domain gap} in literature and is an active field of research in machine learning \cite{BenDavid2007DomainAdaptation, Peng2017VisDA}.

\begin{figure*}[!t]
	\centering
	\includegraphics[width=\textwidth]{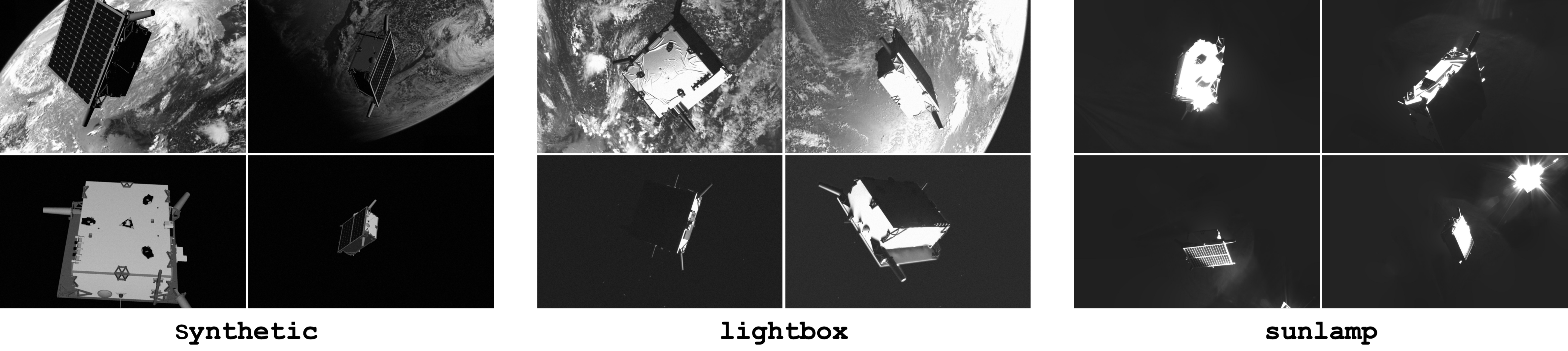}
	\caption{Example images from different domains of SPEED+.}
	\label{fig:SPEED+ overview}
\end{figure*}

While it is possible to train a more domain-invariant CNN model using methods such as domain randomization \cite{Tobin2017DomainRandomization, Zakharov2019DeceptionNet} and texture randomization \cite{Park2019AAS, geirhos2018imagenettrained}, its use in space missions requires a validation of performance in a highly representative space environment prior to mission deployment. Some authors resort to existing spaceborne images captured from previous missions to evaluate the robustness of CNNs \cite{Sharma2019AAS, Park2019AAS, Proenca2019Photorealistic, Black2021PoseEstimationCygnus}. However, for a single target, these images lack diversity in terms of quantity, pose distribution and variability of environmental factors, which severely restrict the comprehensive evaluation of the model's robustness for future missions or for future phases of the same mission. Moreover, these images mostly lack accurate pose labels, which has forced the previous evaluations to be strictly qualitative \mbox{\cite{Proenca2019Photorealistic} or based on hand-labeled pose annotations \mbox{\cite{Black2021PoseEstimationCygnus}}}. Therefore, an alternative approach is to instead physically re-create and simulate the space environment using an on-ground Hardware-In-the-Loop (HIL) robotic testbed which is capable of capturing an arbitrary number of images of the target mockup model and annotating accurate pose labels for each image sample. Then, these HIL images can be used as a weaker surrogate of the target spaceborne images to evaluate on-ground the robustness of a CNN model on an image domain different from the synthetic training images.

To the best of the authors' knowledge, the only publicly available benchmark dataset on spacecraft pose estimation with fully labeled images from multiple sources is the Spacecraft PosE Estimation Dataset (SPEED) \cite{Sharma2019SPEEDonSDR} which consists of 15,300 images of the Tango spacecraft from the PRISMA mission \cite{Damico2014IJSSE, PRISMA_chapter}. Specifically, the dataset comprises 15,000 synthetic images rendered with OpenGL-based Optical Stimulator (OS) camera emulator software \cite{SharmaBeierle2018_CNN, Beierle2019} of the Stanford's Space Rendezvous Laboratory (SLAB) multi-Satellite Software Simulator ($S^3$) \cite{Giralo2018GNSSTestbed} and 300 simulated images captured from the robotic Testbed for Rendezvous and Optical Navigation (TRON) at SLAB. The dataset was made available as part of the Satellite Pose Estimation Competition (SPEC2019) co-hosted by SLAB and the Advanced Concepts Team (ACT) of the European Space Agency (ESA). Throughout the competition, the submitted entries were evaluated and ranked based on their performances on the synthetic test set images. While the models could be evaluated on the 300 simulated images as well, their restricted pose and illumination configurations fall short to perform a comprehensive analysis of the model's robustness across domain gap.

In order to overcome the aforementioned challenges and facilitate the study of ML robustness for future space missions, this paper presents SPEED+, the next-generation spacecraft pose estimation dataset with specific emphasis on bridging the domain gap. As shown in Figure \ref{fig:SPEED+ overview}, SPEED+ comprises large-scale labeled synthetic imagery (\texttt{synthetic}) for training and two unlabeled simulated HIL imageries (\texttt{lightbox}, \texttt{sunlamp}) with distinct visual features and characteristics for testing. More HIL examples are visualized in Figures \ref{fig:lightbox_tangoedges}, \ref{fig:sunlamp_tangoedges} of Appendix \ref{appendix:subsection:projected_examples} for reference. SPEED+ is made publicly available to the aerospace community and beyond as part of the second international Satellite Pose Estimation Competition (SPEC2021). Moreover, a few baseline performance studies are conducted using existing spacecraft pose estimation CNN and domain-bridging algorithms to characterize the domain gap and learnability of SPEED+.

In summary, the contributions of SPEED+ are as follows:
\begin{itemize}
	\item SPEED+ is a first-of-a-kind dataset for vision-only spacecraft pose estimation and relative navigation with emphasis on domain gap between synthetic training images and Hardward-In-the-Loop (HIL) test images from a robotic simulation testbed.
	\item The test images of SPEED+ extend the full orientation space and distance up to 10 m with realistic re-creation of Earth albedo and direct sunlight present in spaceborne imagery. SPEED+ provides unique and unprecedented quantity and quality of mockup spacecraft images, which allow comprehensive evaluation of robustness of spaceborne ML models under a wide range of high-fidelity simulation of environmental settings.
	\item The SPEED+ dataset is used in the second international Satellite Pose Estimation Competition (SPEC2021) with emphasis on robustness of spaceborne ML models across domain gap. The baseline studies in this work exhibit and justify the challenges associated with bridging the performance gap between the synthetic training and HIL test images.
\end{itemize}

\section{Related Work}

\subsection{Datasets for spacecraft pose estimation} 
Apart from SPEED, a number of other datasets for spacecraft pose estimation and spaceborne computer vision problems have been made public as well. For example, URSO \cite{Proenca2019Photorealistic} consists of synthetic and spaceborne images of Soyuz and Dragon spacecraft rendered on Unreal Engine 4, but the labels for spaceborne images are missing. Dung et al.~\cite{Dung2021CVPRDataset} introduced a dataset comprising about 3,000 synthetic and spaceborne images of random satellites with bounding box and segmentation labels. Other authors have also developed their own datasets, such as synthetic images of the Envisat spacecraft rendered with Blender \cite{PasqualettoCassinis2020CNNEKF, Cassinis2021ORGL} and Cygnus spacecraft rendered with Blender and its Cycles rendering engine \cite{Black2021PoseEstimationCygnus}. 

\begin{table*}[!t]
	\caption{SPEED/SPEED+ dataset compositions for different domains and splits.}
	\label{tab:speedplus splits}
	\centering
	\begin{tabular}{c|cc|ccc}
		\toprule
		& \multicolumn{2}{c}{SPEED} &  \multicolumn{3}{c}{SPEED+} \\
		\cmidrule(lr){2-3} \cmidrule(lr){4-6}
		splits & \texttt{synthetic} & \texttt{real} & \texttt{synthetic} & \texttt{lightbox} & \texttt{sunlamp} \\
		\midrule
		Train & 12000 & 5 & 47966 & - & - \\
		Validation & - & - & 11994 & - & - \\
		Test & 2998 & 300 & - & 6740 & 2791 \\
		\bottomrule
	\end{tabular}
\end{table*}

\subsection{Testbeds for spaceborne optical navigation}
Testbeds to simulate vision-based closed-loop navigation and control algorithms in general employ air-bearing platforms on a flat epoxy or granite floor with thrusters and actuators, such as ASTROS at the Georgia Institue of Technology \cite{Tsiotras2014ASTROS}, POSEIDYN at the Naval Postgraduate School \cite{Zappulla2017POSEIDYN}, and M-STAR at the California Institute of Technology \cite{Nakka2018MSTAR}. While these testbeds excel at simulating the spacecraft maneuver commands, their capabilities for creating a large-scale annotated dataset have not been showcased. In an effort to do away with reliance on synthetic imagery, Pasqualetto Cassinis et al.~\cite{Cassinis2021ORGL} have recently used the GRASL facility at ESTEC to generate 100 simulated HIL images of the 1:25 mockup model of the Envisat satellite. However, as the target mockup is mounted onto a static tripod, the configurable pose distribution is severely restricted. To the best of the authors' knowledge, there is no publicly available dataset that consists of more than just hundreds of HIL or spaceborne images of a single target with accurate pose labels. This is not sufficient for a comprehensive evaluation of the robustness of a spaceborne ML model.

\subsection{Datasets for domain gap}
A number of datasets with emphasis on domain gap are available for terrestrial computer vision applications, such as Office-31 \cite{Saenko2010Office31}, Syn2Real \cite{Peng2017VisDA, Peng2018Syn2RealAN}, and DomainNet \cite{Peng2019DomainNet} for various tasks such as classification, object detection and semantic segmentation. Perhaps the application most similar to the spaceborne navigation is the semantic segmentation for autonomous driving, in which the large-scale synthetic images from gaming engines such as GTA V \cite{Richter2016GTAV} or SYNTHIA \cite{Ros2016SYNTHIA} are used for training, whereas real street images available from KITTI \cite{Geiger2012KITTI} or Cityscapes \cite{Cordts2016Cityscapes} are used for testing. Evidently, none of these images contain unique, challenging visual features and characteristics encountered in spaceborne imagery.

\section{SPEED+ Overview}
SPEED+ is the next-generation dataset for spacecraft pose estimation with specific emphasis on the model robustness across the domain gap. While the presented SPEED+ consists of images of the Tango spacecraft from the PRISMA mission \cite{Damico2014IJSSE, PRISMA_chapter}, the spacecraft model can be seamlessly exchanged in the future, and the presented methodology of data collection is general. SPEED+ consists of three different domains of imageries from two distinct sources. The first source is the OpenGL-based Optical Stimulator (OS) \cite{SharmaBeierle2018_CNN, Beierle2019} camera emulator software of the SLAB's multi-Satellite Software Simulator ($S^3$) \cite{Giralo2018GNSSTestbed}, which is used to create \texttt{synthetic} domain comprising 59,960 synthetic images. The labeled \texttt{synthetic} domain is split into 80:20 train/validation sets and is intended to be the main source of training of a CNN model. The second source is the TRON facility at SLAB \cite{Park2021AAS}, which is used to generate two simulated HIL domains with different sources of illumination: \texttt{lightbox} and \texttt{sunlamp}. Specifically, these two domains are constructed using the realistic illumination conditions using light boxes with diffuser plates for albedo simulation and a sun lamp to mimic direct high-intensity homogeneous light from the Sun. Compared to synthetic imagery, they capture corner cases, stray lights, shadowing, and visual effects in general which are not easy to obtain through computer graphics. The \texttt{lightbox} and \texttt{sunlamp} domains are unlabeled and thus intended mainly for testing, representing a typical scenario in developing a spaceborne ML models in which the labeled images from the target space domain are not available prior to deployment. Table \ref{tab:speedplus splits} summarizes the compositions and the splits for different datasets under the old SPEED and SPEED+. It shows that, compared to its previous version, SPEED+ offers much larger quantity of HIL images compared to about just 300 HIL (dubbed \emph{real}) images of SPEED.

SPEED+ is made publicly available under the second Satellite Pose Estimation Competition (SPEC2021) and open for the aerospace community and others to develop and compare the performance of the robust spacecraft pose estimation ML models using labeled synthetic and unlabeled HIL datasets. The detailed plan for hosting, maintaining and licensing the dataset and the competition is provided in Appendix \ref{appendix:accessibility}.

\subsection{On Realism of TRON Simulation}
Much effort has been put into ensuring the realism of the HIL images. First, the mockup model is manufactured via a third-party vendor based on a CAD model. While some details are simplified and omitted, critical model features such as antennae, brackets and external detail features are emulated with tolerances of $\pm 0.0625$ in. The solar cells are printed on high performance automotive grade laminated vinyl, and the bus and main body of the model are covered in simulated flat black MLI thermal blankets.

In order to reproduce realistic space environment, the albedo light boxes and the sun lamp are carefully designed and calibrated to accurately capture the illumination characteristics in space. Specifically, the light boxes \cite{LightBox} are designed and calibrated to provide a uniform maximum radiance of 14 W/m$^{2}$sr, which corresponds to the mean radiance from the Earth for an albedo coefficient of 0.3 \cite{Bhanderi2005Albedo} and a solar irradiance of 1366 W/m$^{2}$ \cite{ASTM2000SolarIrradiance}. The sun lamp, on the other hand, consists of a metal halide arc lamp and a paraboloidal mirror, which together are designed to produce a collimated beam with 1.0 solar constant in space (nominal 1357 W/m$^{2}$) and a spectral response close to 6000 K. The readers are referred to Beierle \cite{Beierle2019} for more details.

\section{Generating SPEED+ Images}
This section first describes the process of collecting and post-processing the HIL images of SPEED+\footnote{Example footage of the data collection process is available at \url{https://youtu.be/kMAvjDW5vX4}}. Then, it ends with the brief description of creating the synthetic imagery of SPEED+.

\begin{figure*}[!t]
	\centering
	\includegraphics[width=\textwidth]{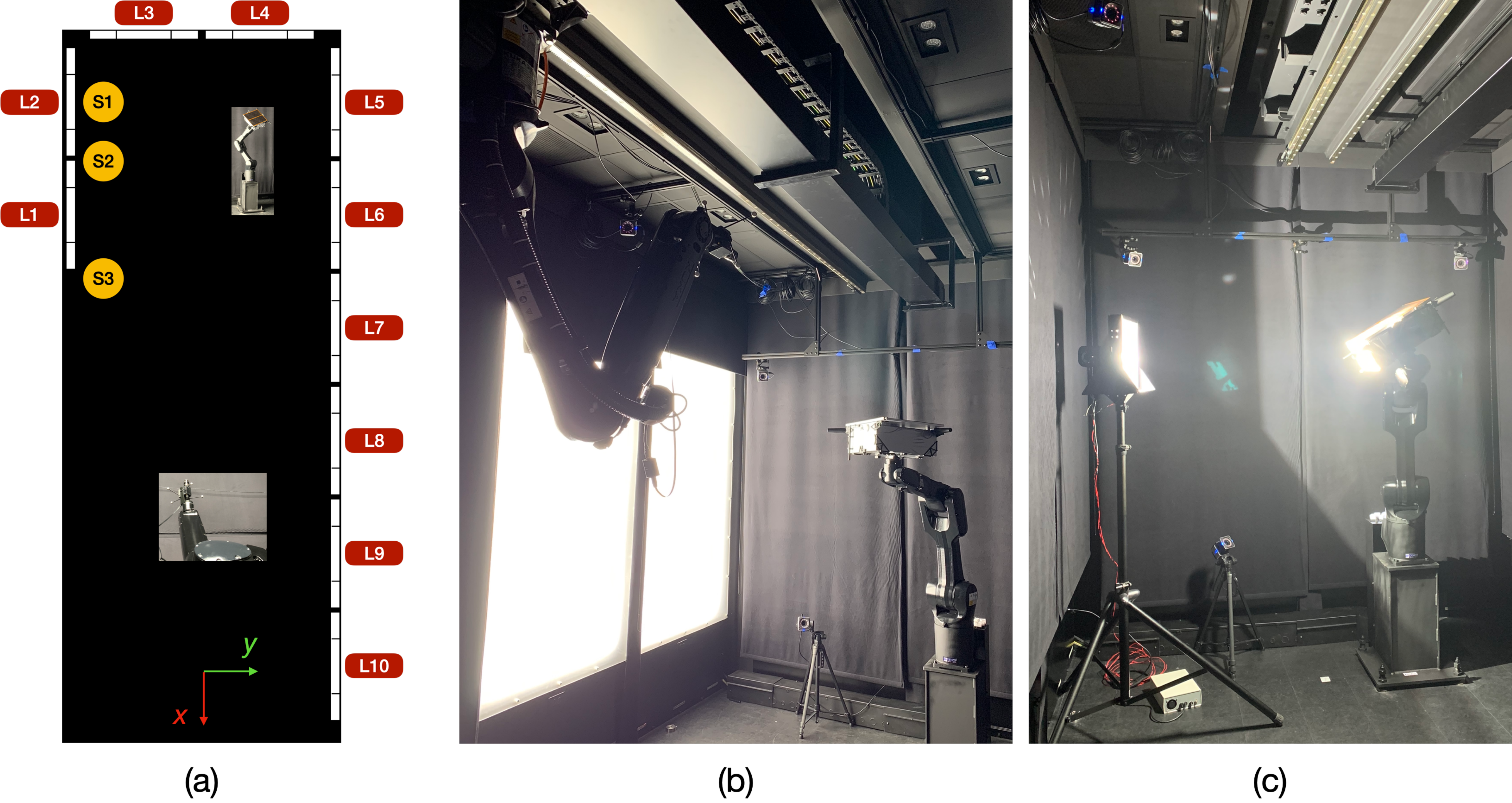}
	\caption{(a) The visualization of the TRON facility simulation room layout. Ten light boxes (L1 $\sim$ L10) and three positions of the sun lamp (S1 $\sim$ S3) are marked and noted. (b) The half-scale Tango spacecraft mockup model is illuminated by two light boxes (L1, L2). (c) The same model is illuminated by the sun lamp placed at S1.}
	\label{fig:TRON visualization}
\end{figure*}

\begin{table*}[!t]
	\caption{Illumination configurations used in SPEED+ simulated imageries.}
	\label{tab:illumination config}
	\centering
	\begin{tabular}{c|cccccccccc}
		\toprule
		\makecell{Illumination \\ Configuration} & 1 & 2 & 3 & 4 & 5 & 6 & 7 & 8 & 9 & 10 \\
		\midrule
		\makecell{Activated \\ Source IDs} & L1 & [L1, L2] & L2 & L5 & [L5, L6] & L6 & [L6, L7] & S1 & S2 & S3 \\
		\bottomrule
	\end{tabular}
\end{table*}

\subsection{TRON Facility Overview}
The TRON facility at SLAB is an 8 $\times$ 3 $\times$ 3 (m) simulation room which consists of two 6 Degrees-Of-Freedom (DOF) KUKA robot arms. One robot arm is fixed to the ground and holds a lightweight, reduced-scale mockup model of the target object, whereas the other arm holds the camera and can move along the ceiling-mounted linear rail across the room. The movements of both objects are tracked using two independent measurement sources: 12 Vicon Vero cameras \cite{Vicon} around the facility which track the infrared (IR) markers attached to both objects, and the KUKA system's internal telemetry of the end-effectors of both robots. The measurements from both sources are used to jointly calibrate the facility, which allows the user to recover the true pose label of the target with respect to the camera from available measurements of an arbitrary sample with millimeter position and sub-degree orientation accuracies at close range. The reader is referred to Park et al.~\cite{Park2021AAS} for details on the facility and its calibration procedure.

In order to emulate a high-fidelity spaceborne illumination setting, the facility contains 10 Earth albedo lightboxes \cite{LightBox} and a metal halide arc lamp. The configuration of the lightboxes is visualized in Figure \ref{fig:TRON visualization}(a) along with the IDs associated with each lightbox (i.e., L1 $\sim$ L10). Figure \ref{fig:TRON visualization}(a) also shows that the sun lamp is placed at three distinct locations at varying angles with respect to the ground-fixed robot arm (S1 $\sim$ S3), with the lamp facing directly towards the spacecraft mockup model. Figures \ref{fig:TRON visualization}(b), (c) show both the lightboxes and the sun lamp in operation. The applicable components of the facility are painted black, and all deactivated sources of illumination are covered with light-absorbing black commando cloths to maximally suppress the reflection and ambient light.

\begin{figure*}[!t]
	\centering
	\includegraphics[width=0.8\textwidth, trim=5 5 5 5,clip]{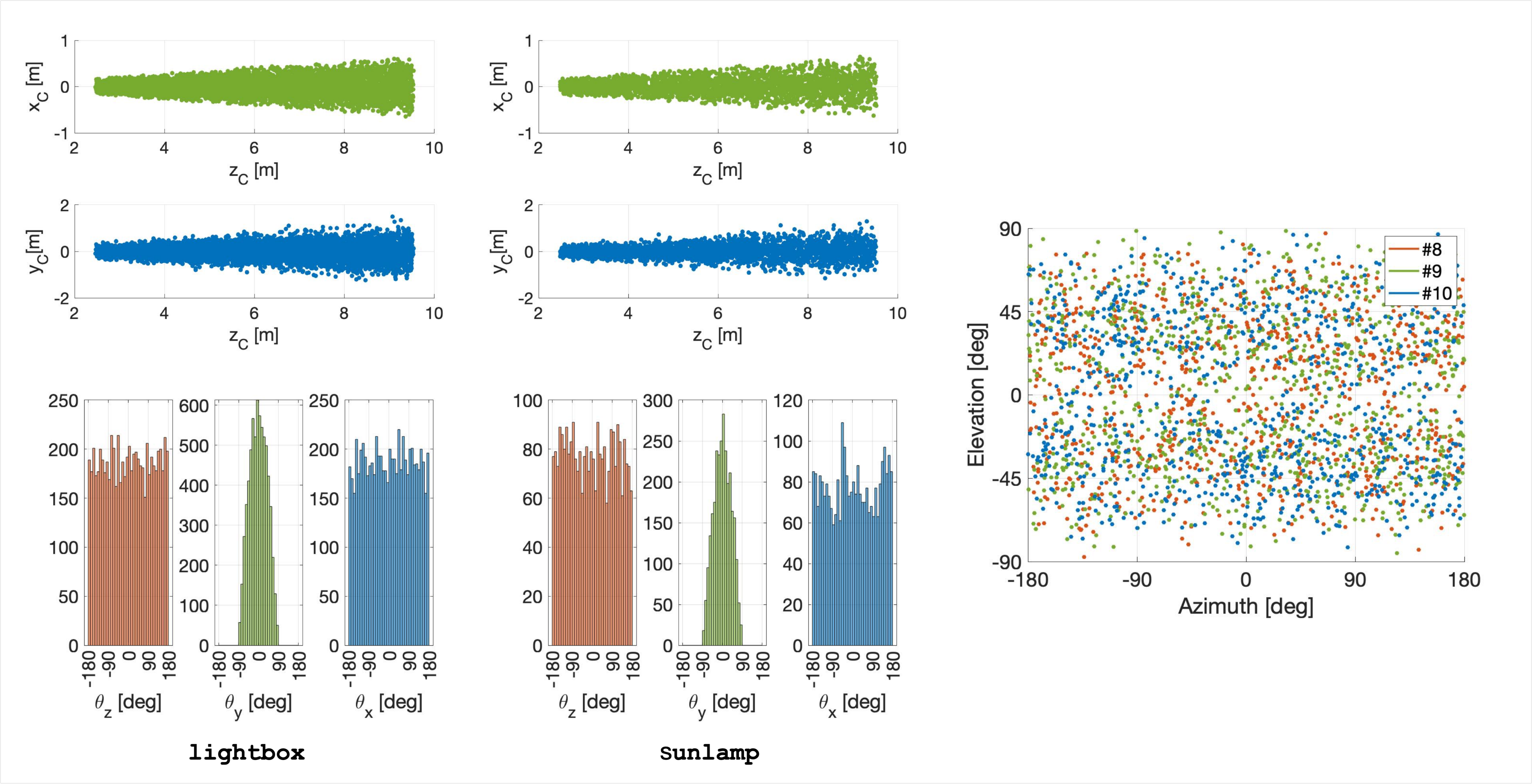}
	\caption{(\emph{Left}) Position (\emph{top}) and orientation (\emph{bottom}) label distributions for \texttt{lightbox}, and \texttt{sunlamp} datasets. The position label is represented in the camera frame ($C$) whose $z$-axis is along the camera boresight, and the $xy$-axes form the image plane. The relative orientation distribution is paremetrized as ZYX Euler angles. (\emph{Right}) The distribution of incident sunlamp direction as spherical coordinates in the target body reference frame ($T$). Samples are color-coded for different illumination configurations for \texttt{sunlamp} domain.}
	\label{fig:label pose distribution}
\end{figure*}

\begin{figure*}[!t]
	\centering
	\includegraphics[width=0.75\textwidth]{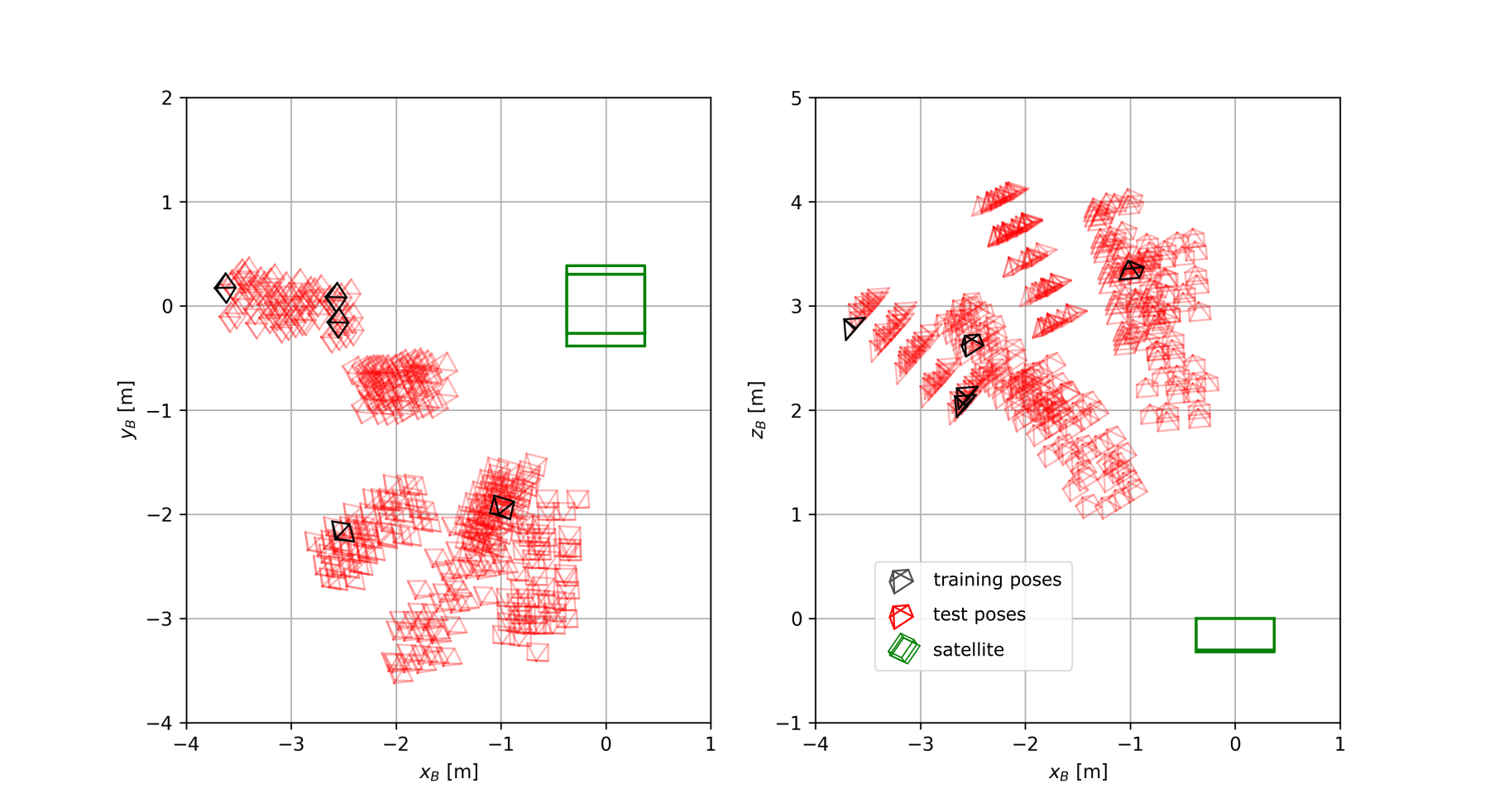}
	\caption{Camera poses for 305 real images of SPEED in the Tango model's body frame ($B$) from two views. The simplified wireframe model of the satellite is plotted in green; camera poses are plotted in red and black for test and training samples, respectively. Directly taken from Kisantal et al.~\protect\cite{Kisantal2020SPEC}.}
	\label{fig:speed real pose distribution}
\end{figure*}

\begin{figure*}[!t]
	\centering
	\includegraphics[width=\textwidth]{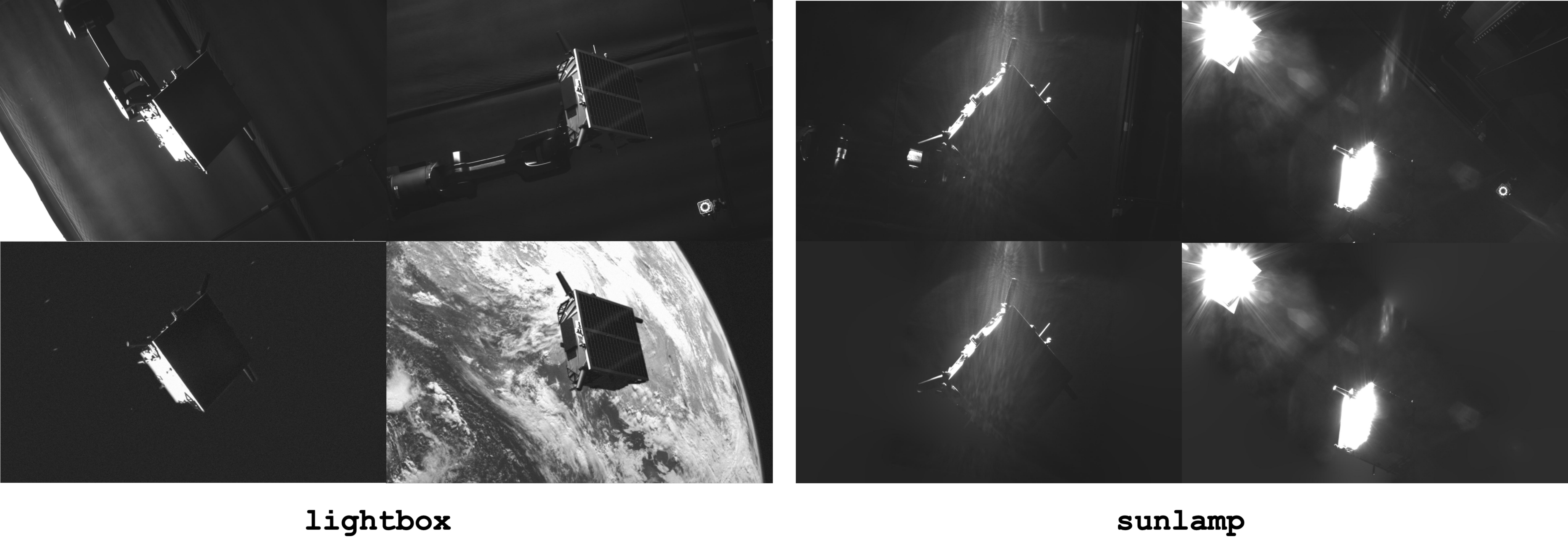}
	\caption{Raw (\emph{top}) and processed (\emph{bottom}) images from the \texttt{lightbox} and \texttt{sunlamp} domains.}
	\label{fig:post-processed examples}
\end{figure*}

\begin{figure*}[!t]
	\centering
	\includegraphics[width=\textwidth]{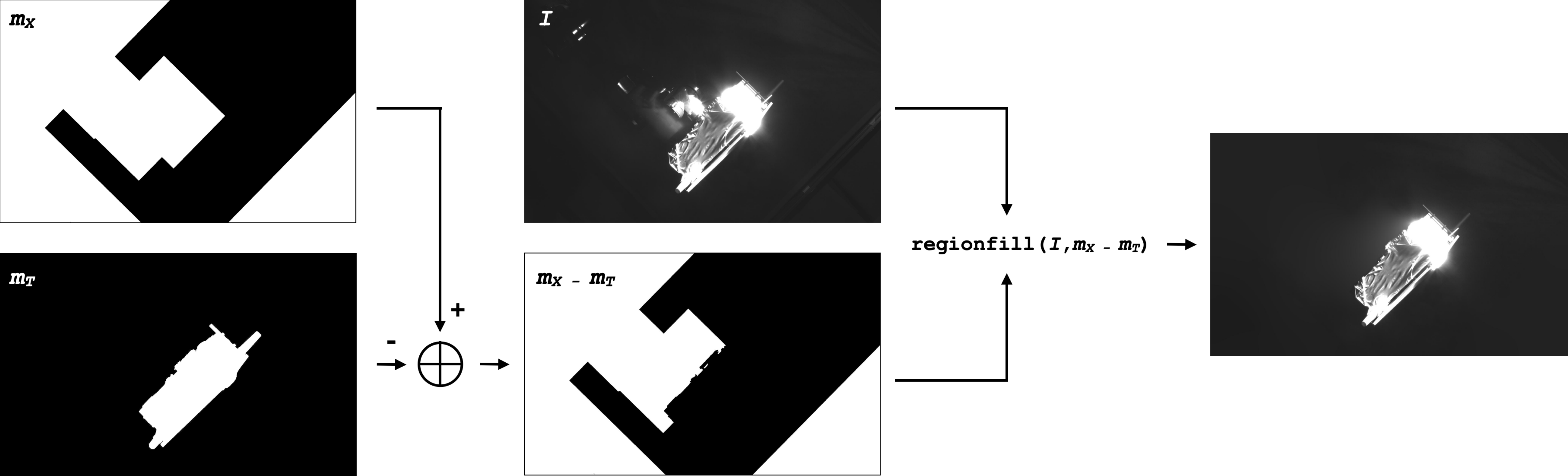}
	\caption{Pipeline for post-processing the \texttt{sunlamp} images.}
	\label{fig:postprocessing sunlamp}
\end{figure*}

\subsection{Collecting SPEED+ HIL Images}
The HIL images of SPEED+ involve a half-scale mockup model of the Tango spacecraft from the PRISMA mission \cite{PRISMA_chapter, Damico2014IJSSE}. The creation of the SPEED+ HIL imageries begins with generating a total of 10,000 pose labels, which are then translated into the commands of both robots. In theory, there is a countless number of pairs of robot commands that will achieve the same prescribed pose label. In order to simplify the process of commands generation and data collection, the target model is commanded at a constant location within the facility regardless of its orientation. Moreover, the camera and the target are commanded such that the camera's boresight is always directed towards the target's location along $-x$-axis as visualized in Figure \ref{fig:TRON visualization}(a). The position labels are generated in compliance with these operational constraints, while ensuring (1) the camera's pose is reconfigurable with the ceiling-mounted robot arm, and (2) the target's center is visible within the image frame of the camera. On the other hand, the orientation labels are simply sampled from the uniform distribution in the \emph{SO}(3) space using the subgroup algorithm \cite{Shoemake1992SubgroupAlg}. The mockup model of the Tango spacecraft is manufactured with two mounting holes at opposite sides, so that depending on the direction of the camera's viewpoint with respect to the model, it can be mounted on either side to ensure the robot arm does not obstruct its view in any way.

With all pose labels translated into the robot arm commands, they are divided into 10 partitions of random 1,000 images based on different illumination configurations shown in Table \ref{tab:illumination config}. It indicates that the configurations 1 $\sim$ 7 employ different sets of light boxes, whereas the configurations 8  $\sim$ 10 employ the sun lamp at different locations along the facility as marked in Figure \ref{fig:TRON visualization}(a). Naturally, the 70\% of the simulated images for configurations 1 $\sim$ 7 are placed under \texttt{lightbox} domain, and the remaining 30\% under \texttt{sunlamp} domain. During the data collection, the intensity of each activated lightboxes is randomly selected from a distribution that is representative of various orbit altitudes for each sample in \texttt{lightbox} domain, whereas the sun lamp is left at constant intensity facing the same direction for \texttt{sunlamp} domain. 

Figure \ref{fig:label pose distribution} depicts the pose distributions of the pose labels for the HIL images, which confirms that for both \texttt{lightbox} and \texttt{sunlamp} datasets, the position and orientation labels are well sampled from the uniform distributions of the 3D Cartesian and \emph{SO}(3) spaces, respectively. This is a big improvement compared to the severely restricted pose distribution of SPEED real images as visualized in Figure \ref{fig:speed real pose distribution}. Moreover, Figure \ref{fig:label pose distribution} shows that for \texttt{sunlamp} domain, the direction of incident light is well-distributed with respect to the mockup target, illuminating it from all possible angles.

\subsection{Post-Processing}

As shown in Figure \ref{fig:post-processed examples}, the raw images include not only the target mockup model of interest but also other components of the facility, such as the robot arm, Vicon cameras, activated lightboxes, etc. While a robust ML model should be able to perform the task of spacecraft pose estimation regardless of the presence of these components, in the setting of the competition, they may pose risks of containing information that can be extracted and exploited to aid the given task. In order to eliminate such possibility altogether, these raw images are post-processed to mask out the unrelated items in the image while simultaneously striving to emulate as closely as possible the visual properties of the spaceborne imagery.

The post-processing procedure for \texttt{lightbox} images loosely follows that of the synthetic imagery of SPEED detailed in \cite{Sharma2019AAS, Sharma2020TAES}. First, the image backgrounds are simply masked out using the binary masks rendered based on the pose labels. Then, either random starfields rendered with the OS software or random patches of the Earth images captured from the Himawari-8 geostationary meteorological satellite \cite{Kotaro2016Himawari} are substitued to the backgrounds, as shown in Figure \ref{fig:post-processed examples}. Specifically, an Earth image with incident sunlight direction that approximately matches the active lightbox configuration is chosen for background substitution. Finally, Gaussian blurring and zero-mean Gaussian white noise are added to emulate the depth of field and additional thermal noise, respectively. Specifically, blurring helps mitigate the sharp boundaries between the satellite foreground and the starfield/Earth background.

On the other hand, the \texttt{sunlamp} raw images contain interesting properties unavailable in \texttt{lightbox} raw images, such as patterned flare introduced by the sun lamp and intense surface glow due to high reflectivity and overexposure of the camera (see Figure \ref{fig:post-processed examples}). In many cases, the glow is too strong such that it obfuscates the shape and boundaries of the target model, effectively acting as an occlusion noise in the image. Therefore, applying the same post-processing as to \texttt{lightbox} images using binary masks will not only eliminate challenging visual effects created by the sun lamp, but it will also give away the exact silhouette of the target model that is otherwise unavailable in the raw images. To mitigate such issues, the post-processing of \texttt{sunlamp} images instead blurs out all the unnecessary components in the image except the general area surrounding the model and the sun lamp, as shown in Figure \ref{fig:post-processed examples}. The overall pipeline is shown in Figure \ref{fig:postprocessing sunlamp}, which involves the approximate mask for all the components that must be removed ($m_X$) and the target model and its surrounding glow effects ($m_T$). Then, given the original image ($I$), the final result is equivalent to the output of the following MATLAB command, \texttt{regionfill($I$, $m_X - m_T$)}, which effectively blurs out the foreground of the given mask.

The final results of the HIL images inevitably include some artifacts that are unique to \texttt{lightbox} and \texttt{sunlamp} images: the infrared markers and the mounting holes, which are not present in the \texttt{synthetic} domain. As these components do not affect the pose labels and are indispensable for the data collection, they are retained in both HIL images. The other artifact unique to \texttt{sunlamp} images is the shape of the surface glow, which is a byproduct of the post-processing step that aims to distinguish the glow effects from the target's surface and the robot arm holding the target. Though somewhat unnatural, the irregular shape of the glow effect is retained as long as it does not add any information about the target's shape. The examples of these artifacts are shown in Figure \ref{fig:artifacts} of Appendix \ref{appendix:subsection:artifacts}.

\subsection{Manual Removal of Samples}
At the end of the post-processing, some samples are manually removed from both \texttt{lightbox} and \texttt{sunlamp} images, leaving total 6,740 and 2,791 images for respective image domains. The majority of rejected \texttt{lightbox} images are due to non-negligible misalignment between the spacecraft and its binary mask when Earth images are inserted as the background. While the pose labels estimated by the facility have millimeter-level and sub-degree-level position and orientation accuracies at close range, it is possible that there may be up to a centimeter-level position error beyond close range due to imperfect calibration of the facility \cite{Park2021AAS}. For example, a position error induced by the misaligned camera boresight would scale linearly with the distance to the target, so 0.1$^\circ$ error would cause 1.75 cm position error at 10 m distance. The misalignment of the binary mask then creates a gap between the spacecraft and the Earth background which could aid the pose estimation task. Therefore, any images with such misalignment observed at more than one edge of the spacecraft are manually discarded by the authors.

For \texttt{sunlamp} images, the samples are rejected based on two criteria. One is the case in which the model's surface reflection is so severe that a human cannot even approximate the pose of the target in an image. The second is when the post-processing step, in an attempt to retain the shape of the surface glow, fails to discriminate and mask the reflection caused by the robot arm holding the model. Some of the rejected samples are visualized in Figure \ref{fig:rejected samples} of Appendix \ref{appendix:subsection:rejectedsamples}.

\subsection{Generating SPEED+ Synthetic images}
Once the outliers are rejected from \texttt{lightbox} and \texttt{sunlamp} domains, the synthetic imagery is created based on the observed pose distribution of the complete HIL imageries. Specifically, the distribution of the position labels encompasses those of the HIL imageries with the separation along $z$-axis sampled from $\mathcal{U}$$(2.25, 10.0)$ [m]. The images are rendered using the camera's intrinsic parameters estimated from the calibration of TRON, which are specified in Appendix \ref{appendix:subsection:cameramodel}. Then, similar to the SPEED synthetic dataset, random Earth backgrounds are inserted to half of the entire \texttt{synthetic} images. Finally, Gaussian blur and noise are added, similar to \texttt{lightbox} images.

\section{Experiments} \label{section:experiments}
This section provides a few baseline performance studies, starting with the \texttt{synthetic}-only experiments (i.e., trained only on \texttt{synthetic} training set) to provide a lower-bound performance on both HIL domains. A number of CNNs with different pose estimation architectures are tested on HIL images to demonstrate the domain gap between the training and test images. Then, two different domain adaptation and randomization algorithms are applied to one of the CNNs as a benchmark. The oracle performances are also approximated by in-domain training (i.e., train and test on the same HIL domain), where the performance is approximated via 5-fold cross validation, i.e., averaged from testing on one of the 5 equal-sized randomly partitioned sets after being trained on the rest. Thus, the oracle represents the upper-bound performance achievable by the given CNN model on the HIL domains. The purpose of these studies is to characterize the domain gap between the synthetic and HIL images and to show that the HIL images and labels from the robotic testbed do retain learnable features. Finally, the same model and the domain randomization algorithm are evaluated on the \texttt{prisma25} dataset, which consists of 25 flight images of the same Tango spacecraft from the PRISMA mission \cite{PRISMA_chapter}. Despite the limited availability of \texttt{prisma25}, the comparative study between SPEED+ HIL and flight images gives insight to the applicability of HIL images as a surrogate for testing.

\subsection{Pose estimation CNNs}
First, the baseline performance study uses three different pose estimation CNNs: the Keypoint Regression Network (KRN) \cite{Park2019AAS} and Spacecraft Pose Network (SPN) \cite{Sharma2019AAS}, which are the two baseline models for SPEED, and HigherHRNet \cite{Cheng2020HigherHRNet}, the state-of-the art model for the bottom-up human joint detection task. These networks employ different strategies for spacecraft pose estimation. For example, SPN simultaneously predicts $N$ closest discrete attitude classes and the relative weights associated with each class, which are then averaged to compute the final attitude. Then, based on the target's model, predicted attitude and bounding box, SPN solves for the relative position by minimizing the residuals between the bounding box and the extremal points of the projected target model using a Gauss-Newton algorithm. On the other hand, KRN directly regress the locations of 11 pre-designated keypoints of the Tango spacecraft's corners and antennae. Then, given the known 2D-3D keypoint correspondences, a Perspective-$n$-Point (P$n$P) problem is solved to output the complete 6D pose \cite{Lepetit2008EPnP}. Finally, HigherHRNet is trained to detect the same 11 keypoints as in KRN, but it returns heatmaps associated with each keypoints. Then, P$n$P is solved for complete 6D pose. 

Note that both KRN and SPN depend on a separate object detection network to detect and crop the bounding box around the spacecraft prior to running the pose estimation networks. In this study, the first-stage object detection network is skipped to simplify the study, especially since the problem of detecting a single known object is a much easier problem than the 6D pose estimation. Instead, the ground-truth bounding boxes computed from the pose labels are used to crop the images around the target during pre-processing. 

\begin{table*}[!t]
	\caption{\texttt{synthetic}-only performances of the baseline models, tested on \texttt{synthetic} validation, \texttt{lightbox} and \texttt{sunlamp} images.}
	\label{tab:results_synthetic_only}
	\centering
	\scriptsize
	\tabcolsep=0.1cm
	\begin{tabular}{c|ccc|ccc|ccc}
		\toprule
		& \multicolumn{3}{c}{\texttt{synthetic}} & \multicolumn{3}{c}{\texttt{lightbox}} &  \multicolumn{3}{c}{\texttt{sunlamp}} \\
		\midrule
		Model & $E_\textrm{T}$ [m] & $E_\textrm{R}$ [$^\circ$] & $E_\textrm{pose}$ [-] & $E_\textrm{T}$ [m] & $E_\textrm{R}$ [$^\circ$] & $E_\textrm{pose}^*$ [-] & $E_\textrm{T}$ [m] & $E_\textrm{R}$ [$^\circ$] & $E_\textrm{pose}^*$ [-] \\
		\midrule
		SPN$^\dagger$ \cite{Sharma2019AAS} & 0.16 & 7.77 & 0.16 & 0.45 & 65.12 & 1.21 & 0.65 & 92.95 & 1.73 \\
		\midrule
		KRN$^\dagger$ \cite{Park2019AAS} & 0.14 & 3.69 & 0.09 & 2.25 & 44.53 & 1.12 & 14.64 & 80.95 & 3.73  \\
		\midrule
		\makecell{HigherHRNet \cite{Cheng2020HigherHRNet} \\ + EP$n$P \cite{Lepetit2008EPnP}}  & 0.05 & 1.51 & 0.04 & 0.97 & 34.71 & 0.77 & 0.85 & 47.75 & 0.98 \\
		\bottomrule
		\multicolumn{10}{l}{\footnotesize ${}^\dagger$ Assuming perfect bounding boxes for cropping during pre-processing.}
	\end{tabular}
\end{table*}

\begin{table*}[!t]
	\caption{Style augmentation \protect\cite{Jackson2019ICCV_StyleAug} and DANN \protect\cite{Ganin2015DANN_ICML, Ganin2015DANN_JMLR} based on KRN \protect\cite{Park2019AAS} tested on \texttt{lightbox}, \texttt{sunlamp} domains of SPEED+ and \texttt{prisma25} datasets. The oracle performances are averaged from 5-fold cross validation on the respective HIL domains.}
	\label{tab:results_domain}
	\centering
	\scriptsize
	\tabcolsep=0.1cm
	\begin{tabular}{c | ccc | ccc | ccc}
		\toprule
		& \multicolumn{3}{c}{\texttt{lightbox}} &  \multicolumn{3}{c}{\texttt{sunlamp}} &  \multicolumn{3}{c}{\texttt{prisma25}}\\
		\midrule
		Method & $E_\textrm{T}$ [m] & $E_\textrm{R}$ [$^\circ$] & $E_\textrm{pose}^*$ [-] & $E_\textrm{T}$ [m] & $E_\textrm{R}$ [$^\circ$] & $E_\textrm{pose}^*$ [-] &  $E_\textrm{T}$ [m] & $E_\textrm{R}$ [$^\circ$] & $E_\textrm{pose}$ [-] \\
		\midrule
		\texttt{synthetic} only &  2.25 & 44.53 & 1.12 & 14.64 & 80.95 & 3.73 & 2.64 & 86.04 & 1.76  \\
		Style Aug.~\cite{Jackson2019ICCV_StyleAug}& 1.06 & 36.14 & 0.81 & 1.32 & 62.85 & 1.32 & 4.06 & 20.57 & 0.71  \\
		DANN \cite{Ganin2015DANN_ICML, Ganin2015DANN_JMLR} & 0.95 & 33.62 & 0.74 & 2.04 & 65.37 & 1.47 & - & - & - \\
		\midrule
		Oracle & 0.24 $\pm$ 0.04 & 6.15 $\pm$ 0.61 & 0.15 $\pm$ 0.01  & 0.19 $\pm$ 0.02 & 5.33 $\pm$ 0.36 & 0.13 $\pm$ 0.01 & - & - & -\\
		\bottomrule
	\end{tabular}
\end{table*}

\subsection{Algorithms for domain gap}
This study adopts two algorithms that represent two distinct approaches to bridging the domain gap: unsupervised domain adaptation \cite{BenDavid2007DomainAdaptation}, which utilizes unlabeled images from the target domain to bridge the feature-level discrepancies, and domain randomization \cite{Tobin2017DomainRandomization}, which instead randomizes various aspects of the input images such that the target image would be considered as another randomized version of the source image. The domain adaptation method used in this study is the Domain-Adversarial Neural Network (DANN) \cite{Ganin2015DANN_ICML, Ganin2015DANN_JMLR}, which employs an adversarial training using a domain classifier to train a domain-invariant feature extractor. For domain randomization, the style augmentation method by Jackson et al.~\cite{Jackson2019ICCV_StyleAug} is used to randomize the spacecraft's texture using the Style Transfer network \cite{Ghiasi2017StyleTransfer}. The style transfer is performed online during the data augmentation phase for half of the training images. More details on the training are presented in Appendix \ref{appendix:training_details}.

\subsection{Metrics}
The pose accuracies are measured using the position error ($E_\textrm{t}$), orientation error ($E_\textrm{q}$), and pose error ($E_\textrm{pose}$) respectively defined as follows,
\begin{align}
	E_\textrm{t} &= \| \tilde{\bm{t}} - \bm{t} \|_2, \\
	E_\textrm{q} &= 2 \arccos | <\tilde{\bm{q}}, \bm{q}> |, \\
	E_\textrm{pose} &= E_\textrm{q} + E_\textrm{t} / \| \bm{t} \|,
\end{align}
where $(\tilde{\bm{t}}, \bm{t})$ respectively denote the predicted and ground-truth relative position vector, $<\tilde{\bm{q}}, \bm{q}>$ denote the inner-product of the predicted and ground-truth quaternion vectors, and the pose error is the sum of the position error normalized by the magnitude of the ground-truth position vector and the orientation error in radians. The pose error, also known as the SPEED score, was the official competition metric used in SPEC2019, chosen as it has shown to properly balance the contribution of the position and orientation errors \cite{Kisantal2020SPEC}. 

In SPEC2021, a modified pose error is used to account for the errors associated with the pose labels of the HIL images. Specifically, from the calibration results of the TRON facility \cite{Park2021AAS}, the following best-performance thresholds are determined: $\theta_q$ = 0.169$^\circ$ for orientation and $\theta_t$ = 2.173 mm/m for normalized translation. Then, if the predicted orientation and translation are both below respective thresholds, the pose error is considered perfect, i.e., for each sample,
\begin{equation}
	E_\textrm{pose}^* = \begin{cases} 0 & \textrm{if}~ E_q < \theta_q ~$\textrm{and}$~ E_t/\||\bm{t}\| < \theta_t \\ E_\textrm{pose} & \textrm{otherwise} \end{cases}.
\end{equation}

\subsection{Results}

The results of the baseline models are shown in Table \ref{tab:results_synthetic_only}. First, it indicates significant domain gap between the synthetic and HIL images, especially for SPN and KRN, as evidenced by the performance drop during testing. The degraded performance on \texttt{lightbox} images indicates that, despite milder illumination condition compared to \texttt{sunlamp} images, the difference in spacecraft's texture between computer graphics and a real, physical model is significant enough to prohibit successful pose prediction. As expected, the performance is much worse when tested on \texttt{sunlamp} images, as they involve much more challenging illumination effects characterized by high contrast and camera overexposure. Interestingly, the performance drop is much smaller for HigherHRNet, which hints that the strategy of heatmap prediction is inherently more robust to domain gap than attitude classification and direct keypoints regression.

\begin{figure}[!t]
	\centering
	\includegraphics[width=0.49\textwidth]{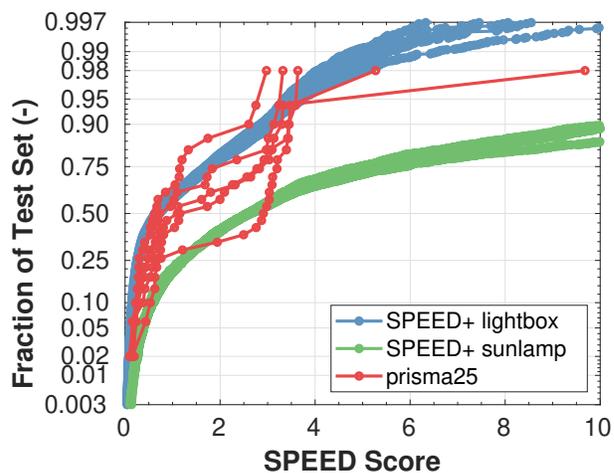}
	\caption{Cumulative distributions of KRN's SPEED score when tested on SPEED+ HIL and PRISMA flight images after 5 training sessions with different random seeds.}
	\label{fig:speed_trend}
\end{figure}

Table \ref{tab:results_domain} shows the performances of KRN after domain-bridging algorithms are applied. First, the oracle performance corroborates the overall accuracy of the pose labels of HIL images, and that they do retain the features learnable by the CNNs to perform the pose estimation task. This serves to justify the use of \texttt{lightbox} and \texttt{sunlamp} images as the test images in SPEC2021. The results of domain adaptation using DANN and domain randomizing using style augmentation showcase the challenges associated with bridging the domain gap between synthetic and HIL images of SPEED+, and that more sophisticated domain-bridging algorithms and hyperparameter tuning will be required to further bridge the gap than a straightforward application of an existing approach. 

Finally, Figure \ref{fig:speed_trend} exhibits the cumulative distribution of KRN's performance on HIL and \texttt{prisma25} test sets after 5 \texttt{synthetic}-only training sessions with different random seeds. Its visualization confirms the challenges associated with \texttt{sunlamp} images compared to those of \texttt{lightbox} for vanilla models trained exclusively on \texttt{synthetic} images. Moreover, despite the limited size of \texttt{prisma25}, Figure \ref{fig:speed_trend} also demonstrates that the distribution curves of SPEED scores on \texttt{prisma25} more or less overlap with those of the SPEED+ HIL images. The coinciding performance trends between SPEED+ HIL and spaceborne images imply a similar level of domain gap against the \texttt{synthetic} domain and the viability of the HIL images as surrogates of the actual flight images in the robustness analysis of CNNs.

\section{Discussions} \label{section:discussions}

This section describes a few limitations and planned future improvements regarding the current SPEED+ and the provided baseline performance studies.

\subsection{Dataset Limitations}
In its currently released form, SPEED+ consists of images of a single known target. However, this paper demonstrates that, given a mockup model of a target, the TRON facility can generate a large number of target HIL images and accurate pose labels with minimal human intervention. Therefore, future sequels of SPEED+ will include more targets to facilitate the study of spaceborne ML techniques beyond the application of pose estimation and navigation about a single known target. Moreover, the KUKA robot arms will undergo a pending update to enable more accurate pose labeling with less human intervention for sample rejection.

\subsection{Relation to Real Mission Constraints and Flight Images}
In real missions, the servicer spacecraft typically does not have access to the target's images until rendezvous. However, in SPEC2021, the unlabeled images from \texttt{lightbox} and \texttt{sunlamp} are available to the public. Such operational constraints are not placed in SPEC2021 to further engage the community's participation: the participants are encouraged to both use and ignore the given unlabeled target images in training robust ML models. 

\subsection{Baseline Studies}
To emphasize, the goal of the provided baseline performance studies is not to showcase the best CNN model architecture and training algorithm for the task at hand, but to characterize and justify the SPEED+ dataset and its suitability for SPEC2021 and robustness studies in spaceborne ML and vision applications. A more comprehensive analysis on the factors contributing to robust ML models will be conducted once the competition concludes.

\section{Conclusions}
This paper presents SPEED+, the next-generation dataset for spacecraft pose estimation and navigation. Compared to its predecessor, SPEED+ focuses on domain gap between training synthetic images and test images captured from a robotic testbed with high-fidelity spaceborne-like illumination conditions. SPEED+ serves to test and compare the robustness of different ML models as part of the international Satellite Pose Estimation Competition that begins in October 2021. This paper introduces and characterizes the SPEED+ dataset using existing pose estimation CNNs and well-established domain adaptation and randomization algorithms.

The presented SPEED+ is an important step toward enabling autonomous proximity operations in space using machine learning techniques to support various future mission concepts, such as refueling defunct space assets and active debris removal. This will help ensure sustained access to near-Earth space, which is a finite resource that needs to be properly understood and managed for the future of mankind.

\acknowledgments
The construction of the testbed was partly funded by the U.S. Air Force Office of Scientific Research (AFOSR) through the Defense University Research Instrumentation Program (DURIP) contract FA9550-18-1-0492, titled High-Fidelity Verification and Validation of Spaceborne Vision Based Navigation. We would like to thank OHB Sweden for the 3D model of the Tango spacecraft used to create the images used in this article and for the flight images in the \texttt{prisma25} dataset.


\appendices{}              
\section{Accessibility} \label{appendix:accessibility}
SPEED+ is made available as part of the second Satellite Pose Estimation Competition (SPEC2021) organized by the Stanford's Space Rendezvous Laboratory (SLAB) and the Advanced Concepts Team (ACT) of the European Space Agency (ESA). The competition begins on October 25th, 2021 and continues till the end of March 2022. This section briefly provides our plan with the hosting, maintenance, and licensing.

\subsection{Kelvins Platform}
SPEC2021 is hosted on the Kelvins platform of ESA, which has hosted the first SPEC (SPEC2019) and many other space-related challenges. The platform includes a link to the dataset, the description of its structure, a leaderboard, and a page for public discussions about the dataset and the competition. More information can be found at \url{kelvins.esa.int/pose-estimation-2021/}.

The SPEED+ dataset is publicly released through the Stanford Digital Repository with a unique DOI \cite{Park2021speedplusSDR}, and the link to the dataset is available on Zenodo as well through the SPEC2021 platform. The dataset is released under the Creative Commons Attribution-NonCommercial-ShareAlike 4.0 (CC BY-NC-SA 4.0) license.

\begin{table*}[t]
	\caption{Calibrated camera parameters}
	\label{tab:calibrated camera}
	\centering
	\begin{tabular}{c|c|c}
		\toprule
		Parameter & Description & Value \\
		\midrule
		$N_u$ & Number of horizontal pixels & 1920 \\
		$N_v$ & Number of vertical pixels & 1200 \\
		$f_x$ & Horizontal focal length [m] & 0.017513 \\
		$f_x$ & Vertical focal length [m] & 0.017513 \\
		$p_x$ & Horizontal pixel length [$\mu$m] & 5.86 \\
		$p_y$ & Vertical pixel length [$\mu$m] & 5.86 \\
		$[r_1, r_2, r_3]$ & Radial distortion parameters & [-0.2238, 0.5141, -0.1312] \\
		$[t_1, t_2]$ & Tangential distortion parameters ($\times 10^{-4}$) & [-6.650, -2.140] \\
		\bottomrule
	\end{tabular}
\end{table*}

\subsection{Baseline Models}
The SPN \cite{Sharma2019AAS} and KRN \cite{Park2019AAS} baseline models, along with DANN \cite{Ganin2015DANN_ICML, Ganin2015DANN_JMLR} and style augmentation \cite{Jackson2019ICCV_StyleAug}, are available in the following GitHub repository: \url{github.com/tpark94/speedplusbaseline}

\section{Camera Model} \label{appendix:subsection:cameramodel}
The SPEED+ HIL images are captured using the Point Grey Grasshopper 3 camera with a Xenoplan 1.4/17mm lens. The camera is calibrated prior to its use, and its calibrated parameters are shown in Table \ref{tab:calibrated camera}.

The radial and tangential distortion parameters follow the conventions used in OpenCV. The specified parameters are provided in the \texttt{camera.json} file.

\section{Training Details} 
\label{appendix:training_details}

This section describes the training details for the baseline studies. All methods are implemented with PyTorch v1.8.0 and trained on an NVIDIA GeForce RTX 2080 Ti 12GB GPU.

\subsection[KRN]{KRN \cite{Park2019AAS}} 
First, the training of KRN \cite{Park2019AAS} largely follows the original work, except it instead uses the AdamW \cite{Loshchilov2017AdamW} optimizer with $\beta_1 = 0.9, \beta_2=0.999$ for backpropagation. When training on \texttt{synthetic} training images, KRN is trained with batch size of 48 for 50 epochs and initial learning rate of 0.001 which decays by the factor of 0.95. When training on the HIL images, the number of training epochs and the decay factor are adjusted according to the number of the available training images.

\subsection[SPN]{SPN \cite{Sharma2019AAS}} 
The implementation of SPN follows the original tensorflow version of the author, which differs from the descriptions in Sharma \& D'Amico \cite{Sharma2019AAS}. In this work, the backbone is the first 5 layers of ImageNet-pretrained AlexNet \cite{Krizhevsky2012AlexNet}, which takes 227 $\times$ 227 images as inputs. The Branches 2 and 3 of SPN, along with the backbone, are trained together simultaneously. The labels are generated from 5,000 attitude classes uniformly sampled from the $SO$(3) space, and at training and inference, 5 neighboring attitude classes are predicted. 

\subsection[HigherHRNet]{HigherHRNet \cite{Cheng2020HigherHRNet}}
The network is directly taken from the official repository\footnote{https://github.com/HRNet/HigherHRNet-Human-Pose-Estimation} and simplified for the application of single-object pose estimation. The exact same data augmentation pipeline for KRN is used here. The input image is resized to 480 $\times$ 320. The keypoint location is taken to be the pixel with the highest intensity in the predicted heatmaps. Then, EP$n$P \cite{Lepetit2008EPnP} is used to solve for the P$n$P problem. In case the HigherHRNet backbone fails to detect enough keypoints to run P$n$P, a dummy solution of $\bm{t} = [0, 0, 5]^\top$ (m), $\bm{q} = [1, 0, 0, 0]^\top$ is returned. Such entries are not skipped to be in agreement with the submission rule of SPEC2021.

\begin{figure}[!t]
	\centering
	\includegraphics[width=0.45\textwidth]{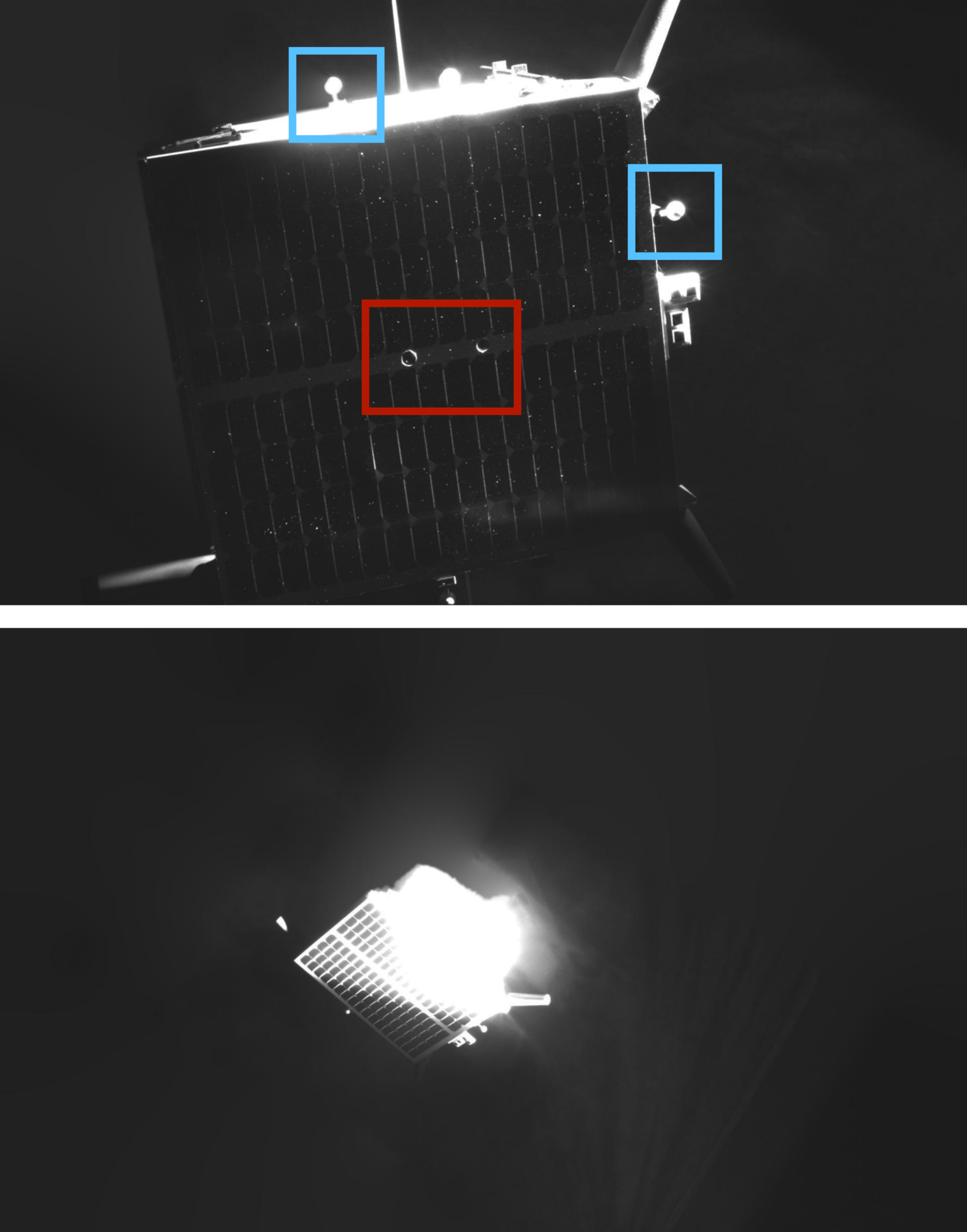}
	\caption{Visualization of the retained artifacts, such as the infrared markers (\emph{top}, blue), mounting holes (\emph{top}, red), and the unnatural shape of the surface glow (\emph{bottom}).}
	\label{fig:artifacts} 
\end{figure}

\begin{figure*}[!t]
	\centering
	\includegraphics[width=0.8\textwidth]{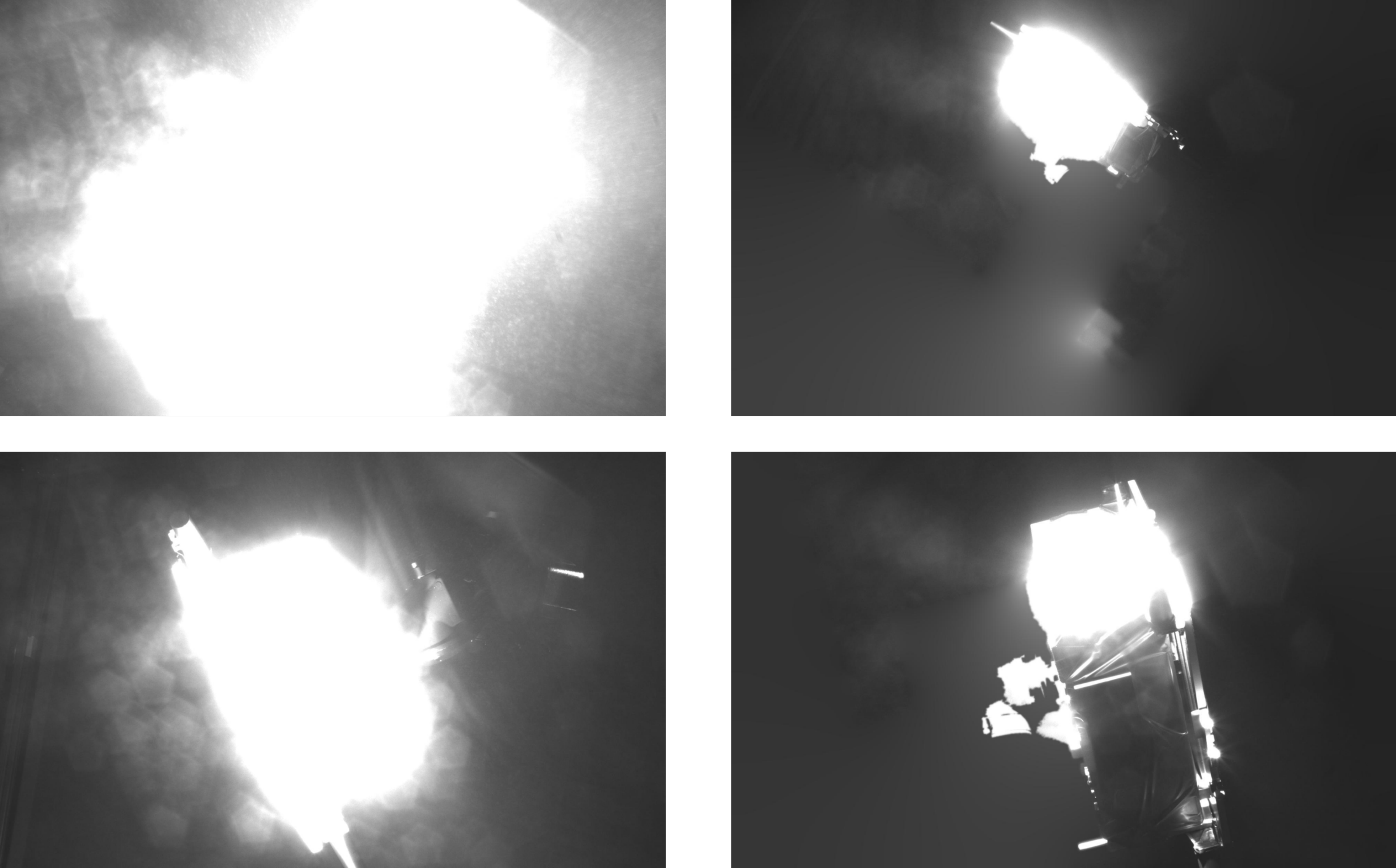}
	\caption{Example samples rejected from \texttt{sunlamp} domain due to extreme surface glow and camera overexposure (\emph{left}) and failed post-processing (\emph{right}).}
	\label{fig:rejected samples}
\end{figure*}

\subsection[DANN]{DANN \cite{Ganin2015DANN_ICML, Ganin2015DANN_JMLR}}
The domain classifier of DANN are attached to the MobileNetv2 \cite{Sandler2018MobileNetv2} feature extractor of KRN, which ends with a feature map of $320 \times 7 \times 7$. Here, after the gradient reversal layer, the domain classifier with two convolution layers is added. Specifically, the first layer is a 1 $\times$ 1 convolution operation to 1280 channels followed by a ReLU activation and an average pooling layer. Then, the second layer uses another 1 $\times$ 1 convolution to end with just 1 channel which is then classified to either source or target domain classes. The training of KRN with DANN follows the training routine of KRN above. 

\subsection[Style Augmentation]{Style augmentation \cite{Jackson2019ICCV_StyleAug}}
The pre-trained style transfer network is directly taken from the official repository\footnote{https://github.com/philipjackson/style-augmentation}. Then, in order to facilitate online stylization at the data augmentation stage, the mean embedding of the SPEED+ \texttt{synthetic} training set is interpolated with a random style embedding from ${\mathbb{R}}^{100}$ with $\alpha$ = 0.5. The training with style transfer network follows the training routine of KRN above.

\section{Visualization} 

\subsection{Retained Artifacts in HIL Images} \label{appendix:subsection:artifacts}

Figure \ref{fig:artifacts} visualizes the retained artifacts unique to HIL images.

\subsection{Images of Rejected Samples} \label{appendix:subsection:rejectedsamples}

Figure \ref{fig:rejected samples} visualizes the samples rejected from the \texttt{sunlamp} domain.

\subsection{Example HIL Images with Model Projection} \label{appendix:subsection:projected_examples}

\begin{figure*}[!p]
	\centering
	\includegraphics[width=0.8\textwidth]{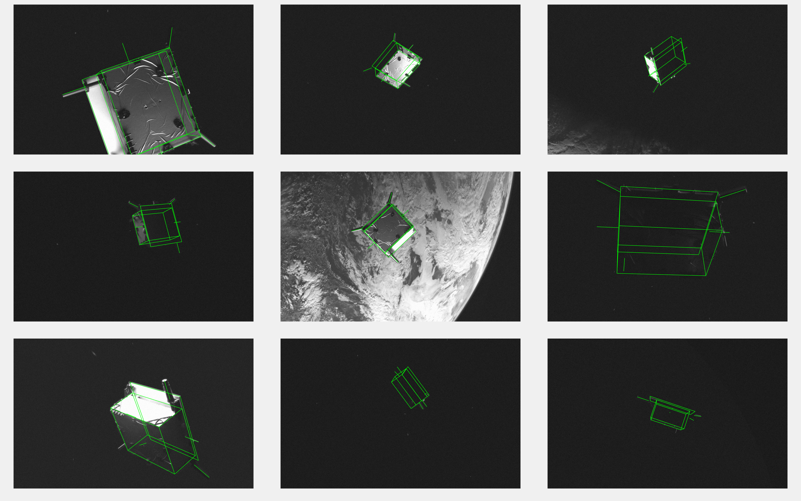}
	\caption{Example \texttt{lightbox} images with the wireframe model of the Tango spacecraft projected based on associated labels and the provided camera properties.}
	\label{fig:lightbox_tangoedges}
\end{figure*}

\begin{figure*}[!p]
	\centering
	\includegraphics[width=0.8\textwidth]{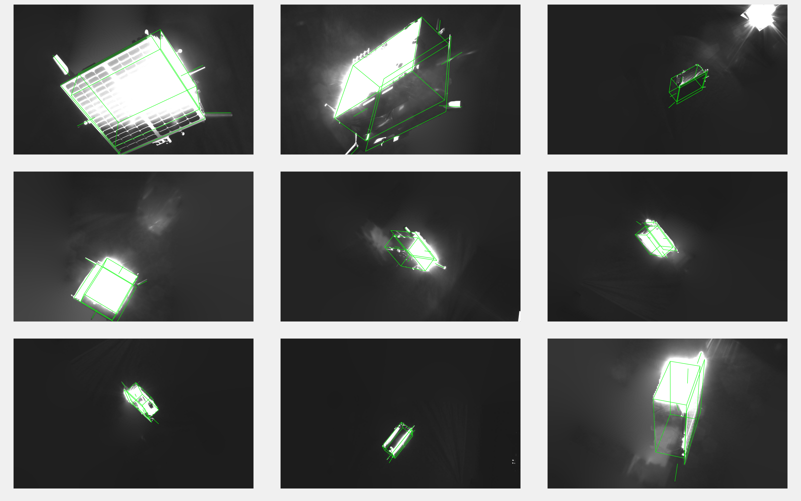}
	\caption{Example \texttt{sunlamp} images with the wireframe model of the Tango spacecraft projected based on associated labels and the provided camera properties.}
	\label{fig:sunlamp_tangoedges}
\end{figure*}

Figures \ref{fig:lightbox_tangoedges} and \ref{fig:sunlamp_tangoedges} show example HIL images from both domains with the wireframe model of the Tango spacecraft projected based on associated labels and the provided camera properties. The alignment between the mockup and the wireframe models validates the accuracy of the HIL image labels estimated from the TRON facility.

\bibliographystyle{IEEEtran}
\bibliography{/Users/taehapark/Documents/reference}

\thebiography

\begin{biographywithpic}
{Tae Ha Park}{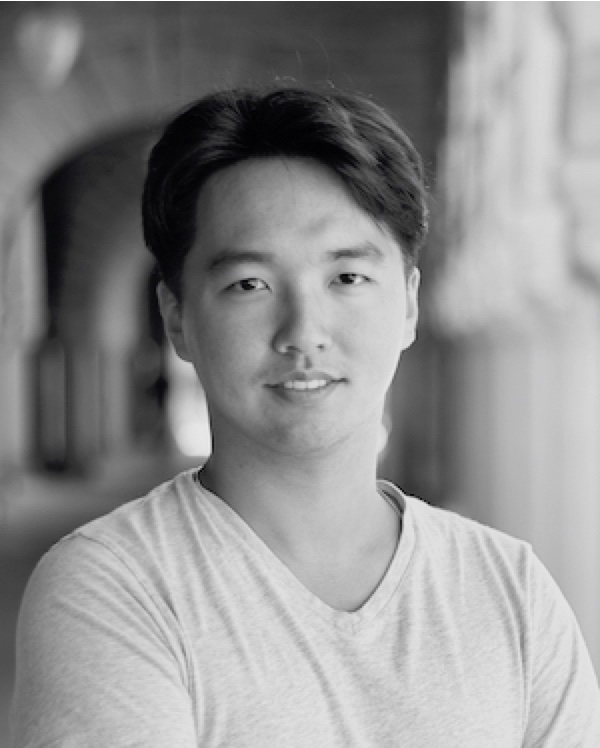}
is a Ph.D.~candidate in the Space Rendezvous Laboratory, Stanford University. He graduated from Harvey Mudd College with a Bachelor of Science degree (2017) in engineering. His research interest is in the development of machine learning techniques and GN\&C algorithms for spaceborne computer vision tasks, specifically on robust and accurate determination of the relative position and attitude of arbitrary resident space objects using monocular vision. Potential applications include space debris removal and refueling of defunct geostationary satellites with unprecedented autonomy and safety measures.
\end{biographywithpic} 

\begin{biographywithpic}
{Marcus M\"artens}{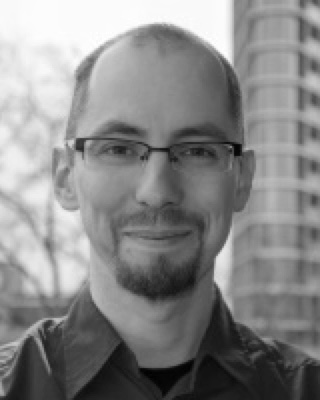}
graduated from the University of Paderborn (Germany) with a Masters degree in computer science. He joined the European Space Agency as a Young Graduate Trainee in artificial intelligence where he worked on multi-objective optimization of spacecraft trajectories. He was part of the winning team of the 8th edition of the Global Trajectory Optimization Competition (GTOC) and received a HUMIES gold medal for developing algorithms achieving human competitive results in trajectory design. The Delft University of Technology awarded him a Ph.D. for his thesis on information propagation in complex networks. After his time at the network architectures and services group in Delft (Netherlands), Marcus rejoined the European Space Agency, where he works as a research follow in the Advanced Concepts Team. While his main focus is on applied artificial intelligence and evolutionary optimization, Marcus has worked together with experts from different fields and authored works related to neuroscience, cyber-security and gaming.
\end{biographywithpic}

\begin{biographywithpic}
{Gurvan Lecuyer}{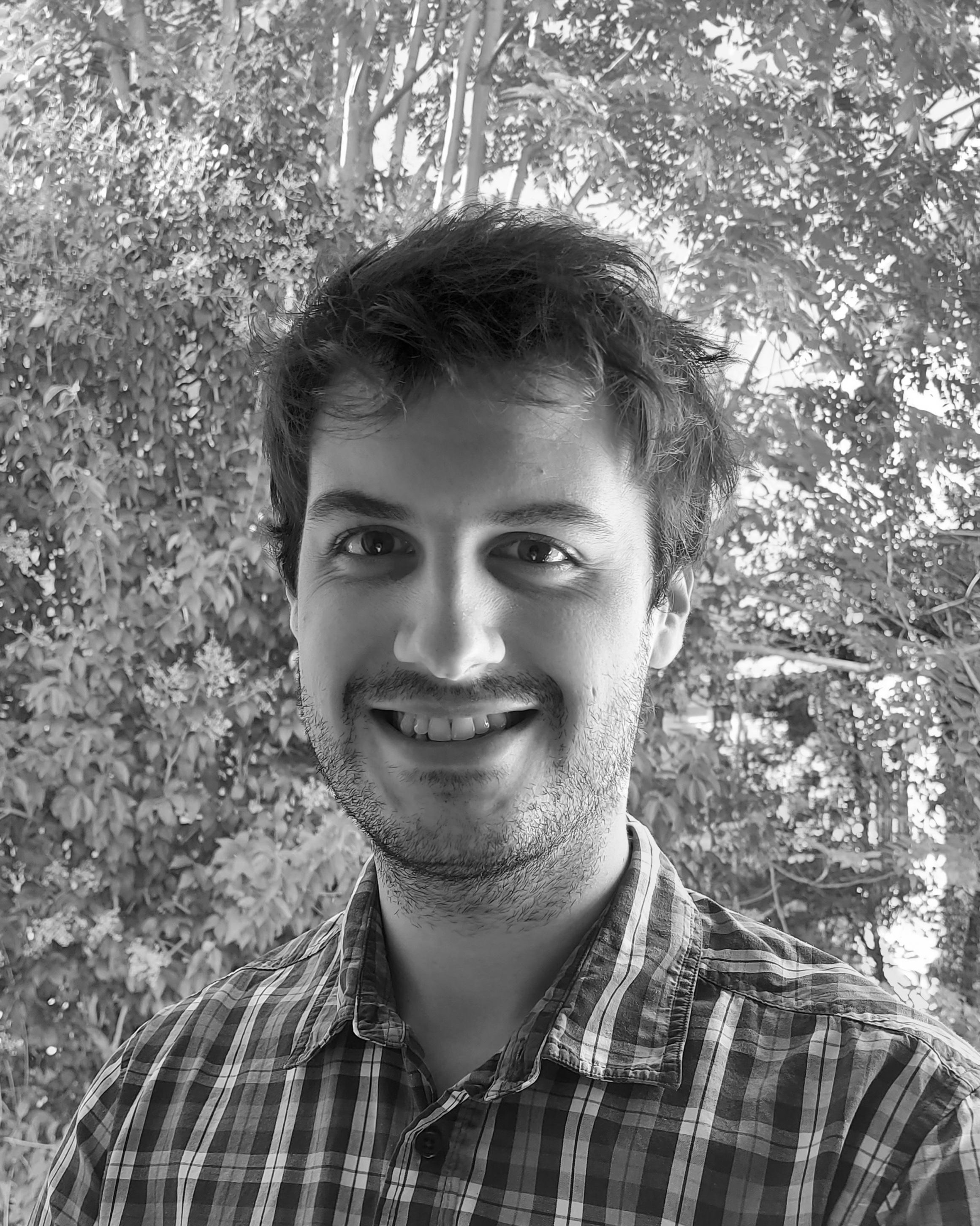}
graduated from the National Institute of Applied Science of Rennes (France) with a Master's degree in Computer Engineering. He then completed a Ph.D.~in signal and image processing at the University of Rennes (France), working on an artificial intelligence algorithm to accelerate the annotation of surgical videos. After that, Gurvan Lecuyer joined the European Space Agency as a Research Fellow in the Advanced Concept Team and $\Phi$-lab to work on artificial intelligence, focusing on the application for Earth Observation.
\end{biographywithpic}

\begin{biographywithpic}
	{Dario Izzo}{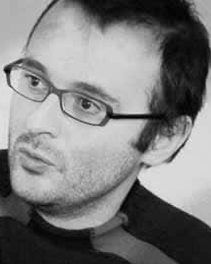}
	graduated as a Doctor of Aeronautical Engineering from the University Sapienza of Rome (Italy). He then took a second master in Satellite Platforms at the University of Cranfield in the United Kingdom and completed his Ph.D.~in Mathematical Modelling at the University Sapienza of Rome where he lectured classical mechanics and space flight mechanics.
	
	Dario Izzo later joined the European Space Agency and became the scientific coordinator of its Advanced Concepts Team. He devised and managed the Global Trajectory Optimization Competitions events, the ESA Summer of Code in Space and the Kelvins innovation and competition platform. He published more than 170 papers in international journals and conferences making key contributions to the understanding of flight mechanics and spacecraft control and pioneering techniques based on evolutionary and machine learning approaches.
	
	Dario Izzo received the Humies Gold Medal and led the team winning the 8$^\textrm{th}$ edition of the Global Trajectory Optimization Competition.
\end{biographywithpic}

\begin{biographywithpic}
	{Simone D'Amico}{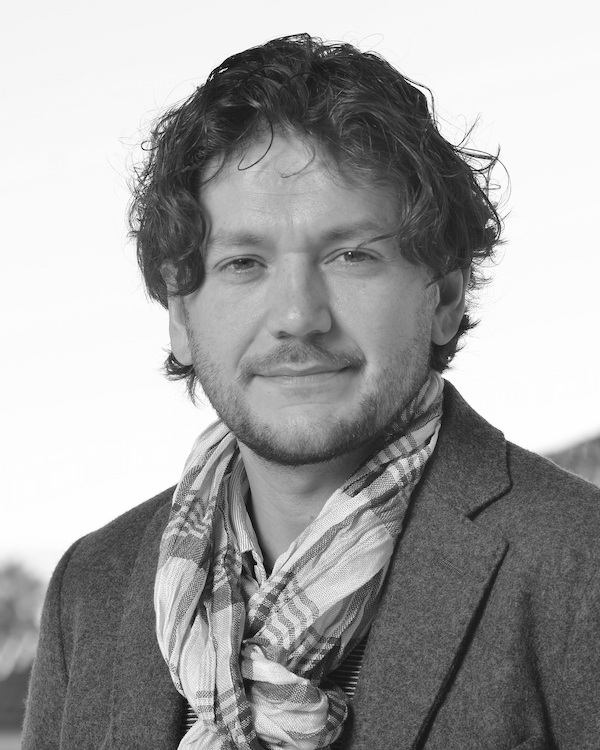}
	received the B.S.~and M.S.~degrees from Politecnico di Milano (2003) and the Ph.D.~degree from Delft University of Technology (2010). From 2003 to 2014, he was research scientist and team leader at the German Aerospace Center (DLR). There, he gave key contributions to the design, development, and operations of spacecraft formation-flying and rendezvous missions such as GRACE (United States/Germany), TanDEM-X (Germany), PRISMA (Sweden/Germany/France), and PROBA-3 (ESA). Since 2014, he has been Assistant Professor of Aeronautics and Astronautics at Stanford University, Founding director of the Space Rendezvous Laboratory (SLAB), and Satellite Advisor of the Student Space Initiative (SSSI), Stanford’s largest undergraduate organization. He has over 150 scientific publications and 2500 google scholar’s citations, including conference proceedings, peer-reviewed journal articles, and book chapters. D'Amico's research aims at enabling future miniature distributed space systems for unprecedented science and exploration. His efforts lie at the intersection of advanced astrodynamics, GN\&C, and space system engineering to meet the tight requirements posed by these novel space architectures. The most recent mission concepts developed by Dr.~D'Amico are a miniaturized distributed occulter/telescope (mDOT) system for direct imaging of exozodiacal dust and exoplanets and the Autonomous Nanosatellite Swarming (ANS) mission for characterization of small celestial bodies. He is Chairman of the NASA's Starshade Science and Technology Working Group (TSWG) and Fellow of the NAE’s US FOE Symposium. D’Amico’s research is supported by NASA, NSF, AFRL, AFOSR, KACST, and Industry. He is member of the advisory board of space startup companies and VC edge funds. He is member of the Space-Flight Mechanics Technical Committee of the AAS, Associate Fellow of AIAA, Associate Editor of the AIAA Journal of Guidance, Control, and Dynamics and the IEEE Transactions of Aerospace and Electronic Systems. Dr.~D’Amico was recipient of the Leonardo 500 Award by the Leonardo Da Vinci Society and ISSNAF (2019), the Stanford’s Introductory Seminar Excellence Award (2019), the FAI/NAA‘s Group Diploma of Honor (2018), the Exemplary System Engineering Doctoral Dissertation Award by the International Honor Society for Systems Engineering OAA (2016), the DLR’s Sabbatical/Forschungssemester in honor of scientific achievements (2012), the DLR’s Wissenschaft Preis in honor of scientific achievements (2006), and the NASA’s Group Achievement Award for the Gravity Recovery and Climate Experiment, GRACE (2004).
\end{biographywithpic}

\end{document}